\documentclass[conference]{IEEEtran}
\IEEEoverridecommandlockouts
\usepackage{cite}
\usepackage{amsmath,amssymb,amsfonts}
\DeclareMathOperator*{\argmax}{arg\,max}
\DeclareMathOperator*{\argmin}{arg\,min}
\usepackage{algorithm}
\usepackage{algorithmic}
\usepackage{graphicx}
\usepackage{textcomp}
\usepackage{xcolor}
\usepackage{dblfloatfix}
\usepackage{booktabs} 
\usepackage[]{subcaption}
\usepackage{xurl}
\urldef{\urlA}\url{https://drive.google.com/file/d/17bwa04yDKuxtE0wPGAL6Mec0HYnPYNfC/view?usp=sharing}
\usepackage{hyperref}

\def\BibTeX{{\rm B\kern-.05em{\sc i\kern-.025em b}\kern-.08em
    T\kern-.1667em\lower.7ex\hbox{E}\kern-.125emX}}
\begin{document}
\raggedbottom
\title{Disentanglement based Active Learning\\
}

\makeatletter
\newcommand{\linebreakand}{%
  \end{@IEEEauthorhalign}
  \hfill\mbox{}\par
  \mbox{}\hfill\begin{@IEEEauthorhalign}
}
\makeatother

\author{\IEEEauthorblockN{\IEEEauthorrefmark{4}Adarsh Kappiyath}
\IEEEauthorblockA{\textit{Flytxt Mobile Solutions} \\
Trivandrum, India \\
kadarsh22@gmail.com}
\and
\IEEEauthorblockN{\IEEEauthorrefmark{4}V S Silpa}
\IEEEauthorblockA{\textit{TCS Research} \\
Pune, India \\
silpavs.43@gmail.com}
\linebreakand
\IEEEauthorblockN{Sumitra S}
\IEEEauthorblockA{\textit{Department of Mathematics} \\
\textit{Indian Institute of Space Science and Technology}\\
Trivandrum, India  \\
sumitra@iist.ac.in}
}

\maketitle
\begingroup\renewcommand\thefootnote{\textsection}
\footnotetext{Equal contribution. Work is completed while authors are at \textit{Indian Institute of Space Science and Technology}, Trivandrum, India.}
\endgroup

\begin{abstract}
We propose Disentanglement based Active Learning (DAL), a new active learning technique based on self-supervision which leverages the concept of disentanglement. Instead of requesting labels from human oracle, our method automatically labels majority of the datapoints, thus drastically reducing the human labeling budget in Generative Adversarial Net (GAN) based active learning approaches. The proposed method uses Information Maximizing Generative Adversarial Nets (InfoGAN) to learn disentangled class category representations. Disagreement between active learner predictions and InfoGAN labels decides if the datapoints need to be human labeled. We also introduce a label correction mechanism which aims to filter out label noise that occurs due to automatic labeling. Results on three benchmark datasets for image classification task demonstrate that our method achieves better performance compared to existing GAN based active learning approaches.
\end{abstract}


\section{Introduction}
A huge amount of labeled data is required to train deep learning models for classification tasks. However, gathering labels for training instances is expensive and time-consuming. Active learning tries to overcome this bottleneck by querying an oracle (for e.g., a human annotator) to label selected unlabeled instances, thus achieving very high accuracy using as few labeled instances as possible\cite{survey}. Pool based active learning chooses active samples from an unlabeled pool using a query strategy based on the active learner. This process is continued until we obtain the required accuracy or reach a fixed labeling budget. Pool based active learning is restricted by the size of the pool. One of the potential methods to overcome this limitation is active learning using query synthesis.

Data generation is widely used for data augmentation in cases where real unlabeled data is scarce. Recent works such as \cite{gaal} and \cite{gen_data} have provided evidence that synthesized data alone can be used to train classifiers that perform as good as classifiers trained on real samples. Owing to the success of Generation based methods, our approach completely relies on generated instances for active learning. A disentangled generative model takes a latent vector as input in which each latent unit controls some semantic aspect in the generated data. For example, a disentangled representation for MNIST dataset consists of each of the generative factors such as digit identity, thickness, width, orientation, stroke length etc., encoded into separate dimensions in the latent space. Disentangled representations have been effective in various downstream tasks like abstract reasoning. 

Information Maximizing Generative Adversarial Nets (InfoGAN) has been proven to be quite successful in disentangling features such as digit identity from digit shapes in MNIST dataset, background images from central digits in SVHN dataset etc.\cite{infogan}. The proposed method utilizes the potential of InfoGAN to disentangle class category representations in an unsupervised fashion. There are several other supervised variants of GANs such as Conditional Generative Adversarial Nets \cite{DBLP:journals/corr/MirzaO14}, Auxiliary Classifier GANs \cite{pmlr-v70-odena17a} which employ label conditioned generation.

In this paper, we introduce Disentanglement based Active Learning (DAL), a new active learning approach based on self-supervision and active labeling. The proposed method aims at reducing human supervision for active learning using disentangled class category representations learned from InfoGAN. The samples for each active learning cycle are synthesized by an InfoGAN out of which some of the datapoints are given to human oracle to label while the others are automatically labeled based on the feedback from the active learner. We also propose a simple yet effective approach to reduce the label noise as a result of automatic labeling which makes our algorithm robust to label noise and reliable.
Our contributions are as follows:
\begin{itemize}
	\item Disentanglement based Active Learning (DAL) is to the best of our knowledge, the first generation based active learning method that uses the concept of disentanglement to decrease the labeling budget in active learning.
	\item DAL achieves the same accuracy as that of the other generation based methods but with a significant reduction in annotation cost.
	\item We perform experimental validation on three benchmark datasets (MNIST, Fashion MNIST, \textit{CIFAR-10}) and extensive ablation studies to prove the effectiveness of our proposed method. 
\end{itemize}

\section{Related Works}
    \textbf{Active Learning.} Active learning is a common approach in many machine learning problems where we have a large pool of unlabeled data but labels are expensive to obtain\cite{survey}. Various criteria are used to select datapoints from the pool, the most common being uncertainty sampling. Our approach is closely related to active learning methods where labels are queried from unreliable and imperfect labelers. Reference \cite{WeakStrong} learns a classifier with low error on data labeled by the oracle while using the weak labeler (oracle) to reduce the overall labeling. Reference \cite{imperfect} and \cite{weak_teachers} investigates how labels obtained from a noisy oracle can be used to improve the performance in active learning.  However, our method is different from the above methods in a way that it counts on the disentangled representations learned by InfoGAN to obtain weak labels, hence eliminating the need for human intervention for the same. Also, DAL majorly focuses on active learning with query synthesis while the above methods are Pool based approaches.

   \textbf{Application of GAN in active learning.} Active learning using GAN (GAAL) is first proposed in \cite{gaal} where uncertain samples are generated by GAN instead of selecting samples from the pool in each iteration of active learning. Reference \cite{QueryLearning} reported poor results due to the synthesis of unrecognizable images. However, \cite{gaal} has shown that active Learning using query synthesis is effective using GANs which synthesizes human interpretable high-dimensional images. Fig. 2 in \cite{gaal} also compares synthesized images from \cite{gaal} and \cite{QueryLearning} which clearly demonstrates the superiority of GAN synthesized instances compared to that of \cite{QueryLearning}. Since our method uses a variant of GAN for image generation, the images in DAL are human interpretable and can be easily annotated.  Our approach generates samples using InfoGAN for active learning but incorporates a mechanism that reduces the burden on human annotators, unlike GAAL. We also provide empirical evidence to show that our method outperforms GAAL.
 Reference \cite{Last2018HumanMachineCF} proposed an active learning method using Conditional GAN (CGAN) specifically for medical image segmentation. There are Pool based active learning approaches which make use of GAN. For instance, \cite{asal} generates the samples based on uncertainty sampling using GAN and nearest neighbors of synthesized  instances from the unlabeled pool are added to the training data in each active learning cycle. Reference \cite{DBLP:conf/iccv/SinhaED19} uses latent space obtained from VAE and an adversarial network to perform Pool based active learning. Reference \cite{BGAD} combines Pool based active learning with data augmentation where the data used to augment is generated from VAE-ACGAN which is as informative as the selected unlabeled instances from the pool in each cycle. Our method exclusively uses generated images for active learning instead of augmenting it with pool images like GAAL.
 
 \textbf{Disentangled Representation learning.} Recently, extensive experiments have been conducted in the area of disentangled representation learning. InfoGAN is a variant of GAN which learns interpretable and disentangled representations in an unsupervised manner\cite{infogan}. There are other variants of GANs such as Structured GAN (SGAN) \cite{NIPS2017_6979} and Triple GAN \cite{NIPS2017_6997} that rely on semi-supervised conditional generation to improve disentanglement. Variational Autoencoders have been extended to learn disentangled representations in \cite{DBLP:conf/iclr/HigginsMPBGBML17}.
 
 \begin{figure*}[t]
\begin{center}
\includegraphics[width=1.0\textwidth]{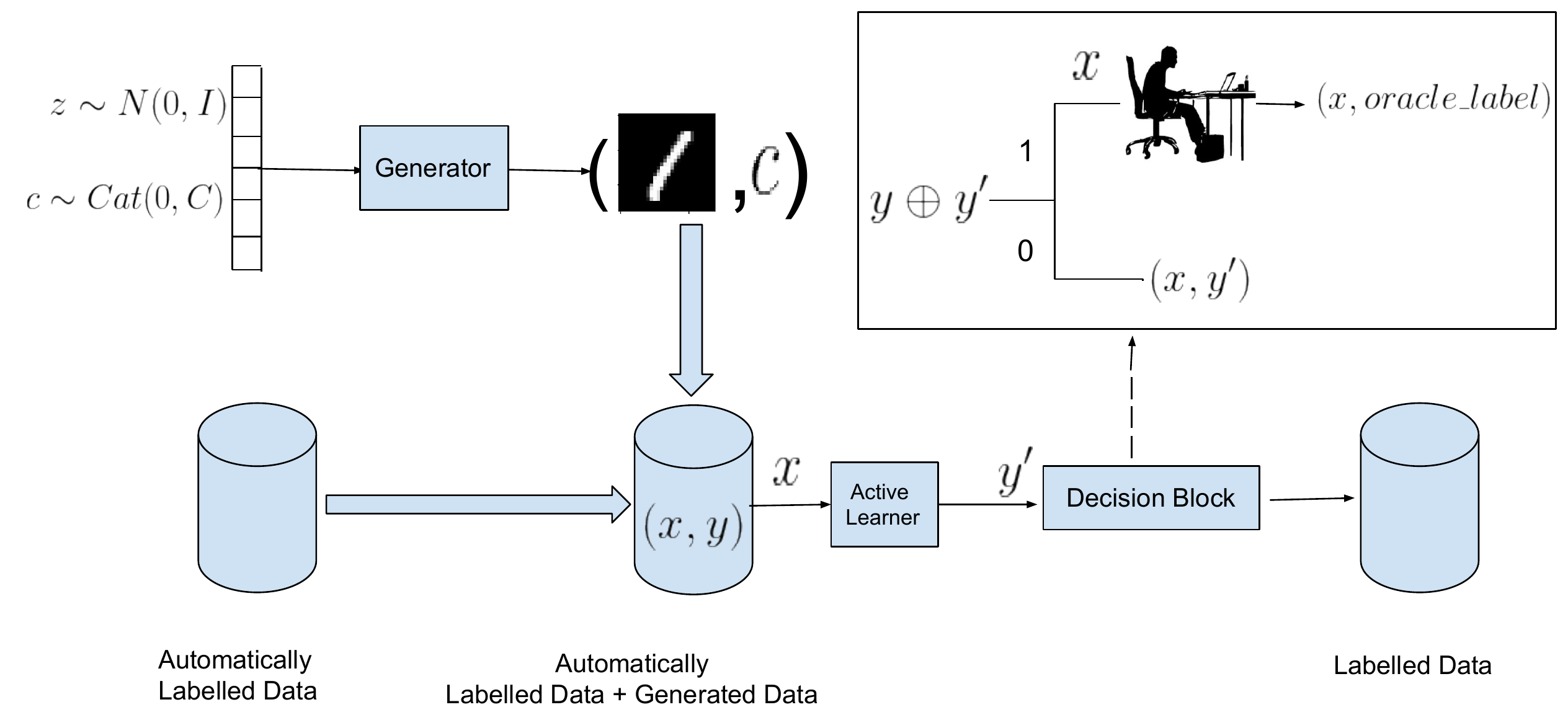}
  \caption{An illustration of DAL. Noise vector and latent code are fed into Generator to generate uncertain instances. They are combined with automatically labeled instances from the previous active learning cycles and are fed into the Active Learner. The decision block compares the active learner predictions with the input latent code for generated images while the comparison is between active learner predictions and previously assigned labels for automatically labeled datapoints. The datapoints are given to a human oracle for labeling in case of disagreement. On the contrary, active learner labels are assigned in case of agreement. The existing labeled dataset is then updated using the newly labeled instances that are used to train the active learner in the subsequent cycle.}
\label{fig:illustration}
\end{center}
\end{figure*}

\section{Proposed Method}
We propose a novel GAN based active learning approach which utilizes the property of disentanglement in InfoGAN. We consider a $C$ class classification problem. Let $D_l=\left \{ (x_i,y_i) \right \}_{i=1}^{n_l}$ and $D_u=\left \{ x_j \right \}_{j=1}^{n_u}$ be the set of labeled and unlabeled instances respectively, where $y_i \in \left \{ 1,2,..,C \right \}$. An illustration of DAL is given in Fig.~\ref{fig:illustration}.
\subsection{Training InfoGAN}
InfoGAN is a GAN variant that learns disentangled representations by maximizing mutual information between generated image and corresponding input latent code \cite{infogan}. InfoGAN trained in a purely unsupervised fashion encodes any of the semantic features (digit identity, thickness, orientation, etc.) in the input latent code. But semi-supervision on class labels guarantees the discrete latent code to capture class categories in the original dataset.
This is because the inductive bias or weak supervision employed is supportive of discovering class categories. Reference \cite{DBLP:conf/aaai/ChenB20} and \cite{ss} utilize weak and semi supervision techniques which ensures that class categories are encoded in the discrete latent codes. Additionally, we also believe that the semi-supervision on class labels guides the discrete latent codes to axis align with the true class categories even though this claim is not theoretically proved in the case of InfoGAN. Hence, once the InfoGAN is trained, we manually ensured that each of the classes corresponds to its original class category for any dataset by generating a few samples from InfoGAN and visually inspecting them. Our experimental results on all three datasets (Fig.~\ref{fig:mnist_gen}, Fig.~\ref{fig:fashion_gen}, and Fig.~\ref{fig:cifar_gen} respectively) demonstrate the same. Owing to these benefits, we use the initial labeled set $D_l$ for supervision during the training of InfoGAN. In the remainder of this paper, we would be referring to the discrete latent codes which capture digit identity as 'latent codes' for simplicity.

We do not use fully supervised variants such as Conditional GANs because we do not have sufficient labeled instances to train these models to achieve comparable performance to that of InfoGAN. To assert this, we provide a quantitative evaluation in Table~\ref{tab:cgan} on how the accuracy of Conditional GAN is affected by the amount of labeled data.

The input to InfoGAN is a noise vector ($z$) sampled from a simple distribution as in ordinary GAN along with latent code $c$ which has meaningful effects on the output. Training of InfoGAN is done as in GANs but also maximizes the mutual information between the latent code and the generator output by minimizing the cross-entropy loss between the input latent code and predictions from Q network. We also train Q network by minimizing the cross-entropy loss between ground truth labels and predictions for $D_l$ from Q to incorporate semi-supervision. This mode of training is similar to restricted labeling in \cite{DBLP:conf/iclr/ShuCKEP20} and semi-supervised InfoGAN as described in \cite{ss}. The generator G can then be used to generate images from required classes upon completion of the training procedure.

\subsection{Disentanglement based Active Learning}
\subsubsection{Generation of uncertain samples using InfoGAN}
Uncertain instances are generated from generator G in each active learning cycle by minimizing the optimization problem as in ASAL\cite{asal}. Uncertainty is computed for each generated sample using information entropy as in \cite{Joshi09n.:multiclass}. We use information entropy because it is suitable for both binary and multiclass classification problems. The samples with maximum entropy are considered as the most informative samples. This can be formulated as an optimization problem which is given by 

\begin{equation}
    \label{eq:max-ent-gan}
    z^{*} =  \argmin_{z} \sum_{q\in \mathcal{C}} p_{\theta(k)}(q|G(z,c)) \log p_{\theta(k)}(q|G(z,c))
\end{equation}

 where the active learner at cycle $k$ is parameterized by $\theta_k$. Once we get $z^{*}$, uncertain samples are generated by passing it through the generator $G$ of InfoGAN.

\subsubsection{Labeling criteria}
The uncertain sample obtained, $x_g = G(z^*,c)$ is labeled either automatically or by an oracle. Ideally, all the generated samples can be added to the training set for the next active learning cycle as in GAAL since the class information is already obtained from latent code. However, InfoGAN can be erroneous in disentangling class categories.

The key idea in our approach to alleviate this problem is to use the knowledge acquired by the active learner till the current active learning cycle. The uncertain instances are fed to active learner to obtain the predictions. The disagreement between active learner predictions and corresponding input latent code acts as a labeling criterion. DAL automatically labels the generated instances in case of consensus between active learner and InfoGAN while the other instances are given to an oracle to query its labels. 

Let $x_g = G(z^*,c)$ be generated at active learning cycle $k$ and $\hat{c}$ be the class corresponding to one-hot latent code $c$, then

\begin{equation}
        \label{eq:label_proc}
         D_{l}\leftarrow
         \begin{cases}
             D_{l}\cup \left \{ x_g,\hat{c} \right \},& \text{if } \underset{q}\argmax\  p_{\theta_{k}}(q|x_g) == \hat{c}\\
            D_l \cup \left \{ x_g,y_g \right \},              & \text{otherwise}
         \end{cases}
\end{equation}

where $y_g$ is the label provided by an oracle for the generated image $x_g$.

But there can be cases where both the latent code and prediction of active learner agree but the generated image could be from a different class. This would lead to incorrectly labeled datapoints (we refer to this as noise) being added to the training dataset. Accumulation of such noise in the training dataset would result in a sub-optimal solution since deep learning methods are negatively affected by label noise, especially in low data regime. We propose a simple label correction technique to diminish the effect of noise on our approach.

\subsubsection{Label correction}
Let $k_i$ be any active learning cycle and $D_{A}$ be the set of all automatically labeled instances from the previous cycles in the dataset $D_l$. As discussed above, a subset of $D_{A}$ contributes to noise in $D_l$ at cycle $k_i$. Hence, we compute active learner predictions for all instances in set $D_A$ in the active learning cycles, $k>k_i$ , since the active learner gets better in each cycle and it learns to classify the noisy instances in $D_A$ to correct class. This leads to disagreement between the noisy labels and predictions of the active learner at some cycle $k>k_i$, which is then provided to the oracle for labeling. This label correction mechanism ensures that the majority of the noisy instances added to $D_l$ at any cycle $k_i$ are added back to $D_l$ as noise-free instances at cycle $k>k_i$, thus reducing the effect of noise on our system due to self-supervision.

\begin{algorithm}
	\caption{Disentanglement Based Active Learning (DAL)}
	\label{alg:loop}
	\begin{algorithmic}[1]
	    \REQUIRE A labeled set ${D_l} =  \left \{ (x_i,y_i) \right \}_{i=1}^{n_l} $ and an unlabeled set  ${D_u} = \left \{ x_j \right \}_{j=1}^{n_u}$ 
	    \REQUIRE  $m$ : Number of datapoints added to the training dataset in each active learning cycle.
		\STATE Train InfoGAN on unlabeled dataset $D_u$ along with supervision provided by the labeled dataset $D_l$.
		\REPEAT
		\STATE Train Active learner on training dataset ${D_l}$.
		\STATE Generate a set of $m$ datapoints, $D_G = \left \{ x_{g_{i}} \right \}_{i=1}^{m}$ where $x_{g_{i}}$ is generated using $z^*$ obtained from \eqref{eq:max-ent-gan}.
		\STATE Update $D_G$ by concatenating $D_G$ with $D_A$, where $D_A$ consists of instances in $D_l$ that are automatically labeled in the previous active learning cycles.
		\STATE For each datapoint in $D_G$, determine how it needs to be labeled using \eqref{eq:label_proc}.
	    \STATE Update $D_l$ by adding labeled instances from $D_G$.
		\UNTIL{Labeling budget is reached}.
	\end{algorithmic}
\end{algorithm}

\begin{figure*}[!b]
\begin{center}
\includegraphics[width=.3\textwidth]{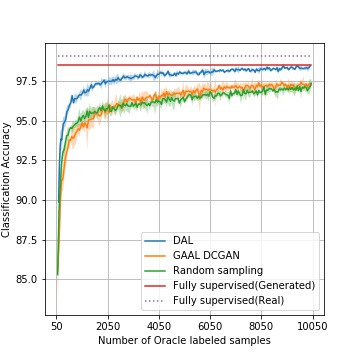}
\includegraphics[width=.3\textwidth]{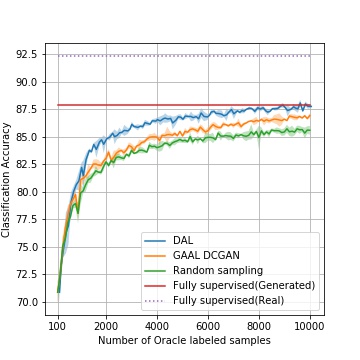}
\includegraphics[width=.3\textwidth]{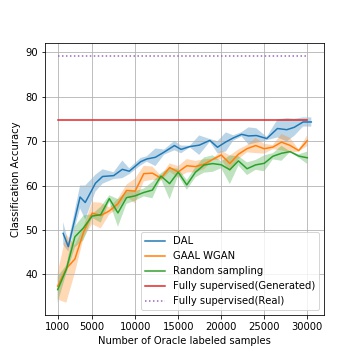}
  \caption{Classification performance of DAL, (Left) MNIST, (Middle) F-MNIST and (Right) \textit{CIFAR-10} dataset.}
\label{fig:baseline_results}
\end{center}
\end{figure*}

\section{Experiments}
We conduct experiments on three different datasets, MNIST \cite{lecun2010mnist}, Fashion MNIST \cite{xiao2017/online} and \textit{CIFAR-10} \cite{cifar10}. We also implement and compare our approach to four different schemes that are listed as follows:
\begin{itemize}
    \item GAN based Active learning (GAAL) from \cite{gaal}.
    \item Random sampling using GAN, where random instances synthesized by GAN are provided to the oracle to label.
    \item Fully supervised learning with generated images, where images generated from GAN are annotated by an oracle and are used to train the classifier. Number of generated instances is same as that of the size of pool. Test accuracy is computed on real testing data distribution.
    \item Fully supervised learning, where all samples from real data are used to train and test the classifier.
\end{itemize}

The architecture of InfoGAN used for the proposed approach is described in \cite{infogan}. We use all the unlabeled instances in the pool along with the initial labeled dataset $S$ to train InfoGAN so that it learns better disentangled representations for class categories. The base architecture used for InfoGAN is that of DCGAN \cite{dcgan} for MNIST and Fashion MNIST while InfoGAN is combined with WGAN \cite{wgan} for \textit{CIFAR-10}.
All the baseline methods are implemented using DCGAN for MNIST and Fashion MNIST while WGAN is used for \textit{CIFAR-10} dataset for a fair comparison with our method. Better models such as Structured GANs \cite{NIPS2017_6979} also improve disentanglement score compared to InfoGAN. However, our aim is to investigate how the concept of disentanglement can be combined with active learning to reduce the labeling budget. We show evidence in our ablation study that models with better disentanglement improve our result.

The authors of GAAL\cite{gaal} have only reported results for binary classification using the distance to the hyperplane as the active-learning criteria. We extend it to a multi-class classification problem using Maximum Entropy as the acquisition function with CNN being the active learner for a fair comparison with DAL.  We use Lenet\cite{Lecun98gradient-basedlearning}, 4-layer CNN and VGG-13\cite{simonyan2014convolutional} as active learners for MNIST, Fashion MNIST and \textit{CIFAR-10} respectively. The architecture details of the active learner and hyperparameters used for all three datasets are given in the  Supplementary material\footnote{See \urlA}. Early stopping is employed to prevent over-fitting. The maximum number of epochs is set to 100. Stochastic gradient descent is used to optimize~\eqref{eq:max-ent-gan}. The gradient update is done for a batch of noise vectors in each active learning cycle instead of minimizing each $z$ vector in a sequential manner which helps to optimize the computation. Minimization is done for 100 gradient update steps where the learning rate is set to 0.001. We initialize the training data with 100 datapoints for MNIST and Fashion MNIST while for a slightly complicated dataset like \textit{CIFAR-10}, we start with 1000 data-points. 50, 100 and 1000 samples are added to the training data in each active learning cycle for MNIST, Fashion MNIST, and CIFAR 10 respectively. This process is repeated until the labeling budget is reached. All experiments are performed for 5 different random seeds. The classification accuracies for all the above schemes are reported on the original testing data.

\begin{table*}[!b]
\caption{Per-Class Generation Accuracies of MNIST, F-MNIST and CIFAR-10 on corresponding GANs trained.}
\centering
\begin{subtable}[t]{0.3\linewidth}\centering
\caption{MNIST using InfoGAN}
 \label{tab:mnist per class gen}
  \begin{tabular}{|p{1.9cm}|p{1.9cm}|}
  \hline
  Class & Accuracy\\
  \hline\hline
    Zero &0.996 \\
    One  &1.0 \\
    Two  &0.982 \\
    Three &0.988 \\
    Four  &0.97 \\
    Five &0.945 \\
    Six &0.993\\
    Seven &0.99\\
    Eight &0.988\\
    Nine  &0.994\\
  \hline
\end{tabular}

\end{subtable}\hfill%
\begin{subtable}[t]{0.3\linewidth}\centering
\caption{Fashion MNIST using InfoGAN}
 \label{tab:fashion_mnist per class gen}
  \begin{tabular}{|p{1.9cm}|p{1.9cm}|}
  \hline
  Class & Accuracy \\
  \hline\hline                   
    T-shirt &0.785 \\
    Trouser  &0.998 \\
    Pull-over  &0.376 \\
    Dress &0.821 \\
    Coat  &0.397 \\
    Sandal &0.728 \\
    Shirt &0.185\\
    Sneaker &0.715\\
    Bag &0.994\\
    AnkleBoot  &0.974\\
    \hline
  \end{tabular}
 
\end{subtable}\hfill%
\begin{subtable}[t]{0.3\linewidth}\centering
 \caption{\textit{CIFAR-10} using InfoWGAN}
  \label{tab:cifar10_per_class_gen_acc}
  \begin{tabular}{|p{1.9cm}|p{1.9cm}|}
  \hline
  Class & Accuracy \\
  \hline\hline                 
    Aeroplane &0.819 \\
    Automobile  &0.868 \\
    Bird  &0.535 \\
    Cat &0.583 \\
    Deer  &0.591 \\
    Dog &0.523 \\
    Frog &0.889\\
    Horse &0.866\\
    Ship &0.819\\
    Truck  &0.858\\
    \hline
  \end{tabular}
 
\end{subtable}
\end{table*}

\begin{figure*}[t]
    \begin{subfigure}[t]{0.3\textwidth}
      \includegraphics[width=\textwidth]{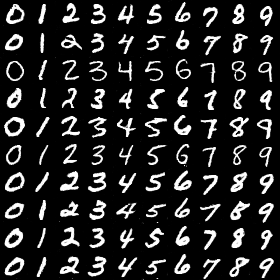}
      \caption{Generated samples from MNIST}
      \label{fig:mnist_gen}
    \end{subfigure}
    \hfill
    \begin{subfigure}[t]{0.3\textwidth}
       \captionsetup{justification=centering}
      \includegraphics[width=\textwidth]{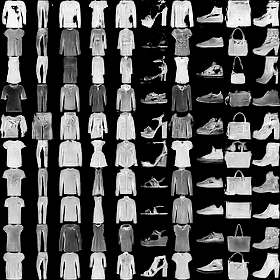}
      \caption{Generated samples from Fashion MNIST}
      \label{fig:fashion_gen}
    \end{subfigure}
    \hfill
    \begin{subfigure}[t]{0.3\textwidth}
      \includegraphics[width=\textwidth]{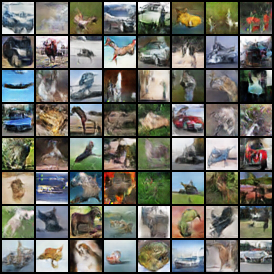}
      \caption{Generated samples from \textit{CIFAR10}}
      \label{fig:cifar_gen}
    \end{subfigure}
    
    \caption{Generated samples from each of the three datasets. Images from MNIST and Fashion MNIST are generated using InfoGAN while InfoWGAN is used to generate images from \textit{CIFAR10}}
  \end{figure*}

\section{Results}
We perform experiments to analyze the classification accuracies and average labeling budget required for different active learning strategies listed above and compare it with the proposed approach. 

 DAL either decides to provide instances to an oracle to label or automatically label them at any active learning cycle. We have included a plot of the validation accuracy versus the number of human labeled datapoints at each cycle in Fig.~\ref{fig:baseline_results}, since our aim is to analyze the reduction of human labeling compared to the other approaches. It is clear from Fig.~\ref{fig:baseline_results} that our method outperforms all other generation based active learning baselines on all three datasets. 
 
 It can be observed that the test accuracy of a classifier trained on the generated images and tested on the real data is less compared to a fully supervised classifier trained on real data. The decrease in performance can be attributed to the GANs used. The performance of any GAN based active learning approaches are limited to the GANs used to generate images and classifier trained on it. For instance, we have used InfoGAN with the base architecture being DCGAN and all other baseline methods are compared to approaches using DCGAN for both MNIST and Fashion MNIST. Performance of classifier trained on images generated from DCGAN is lesser compared to a classifier trained on real images since DCGAN is unable to perfectly learn the real data distribution \cite{Shmelkov18}. However, this problem can be eliminated if classifiers are trained on images generated using better training techniques as shown in \cite{gen_data} which aims to minimize the divergence between generated and real data distribution. Using such techniques would only enhance the results reported and the performance of all the GAN based active learning settings approaches to fully supervised accuracy on the real data distribution.
 
Once the InfoGAN is trained, we compute the generation accuracy to measure the degree of disentanglement of the class category. Generation accuracy is calculated by manually matching the discrete latent code to the actual class for all the generated images. Table~\ref{tab:mnist per class gen}, Table~\ref{tab:fashion_mnist per class gen} and Table~\ref{tab:cifar10_per_class_gen_acc}  shows the per-class generation accuracies for MNIST, Fashion MNIST and \textit{CIFAR-10} respectively. InfoGAN trained on MNIST has a high degree of disentanglement for any class owing to the simplicity of the dataset. Per-class generation accuracies for \textit{CIFAR-10} reveals that InfoWGAN trained on \textit{CIFAR-10} dataset learns to moderately disentangle the classes though some classes such as Bird, Cat, Deer, and Frog have less per-class generation accuracies(53.5\%, 58.3\%, 59.1\%, and 52.3\% respectively). It is interesting to note that the InfoGAN used for Fashion MNIST struggles to disentangle highly correlated classes such as Pull-over, Coat, and Shirt with per-class accuracies 37.6\%, 39.7\%, and 18.5\% respectively. However, in case of varying disentanglement, the active learner in DAL manages to query the oracle for samples from the classes where InfoGAN can't be relied upon while samples from the other classes are automatically labeled, resulting in the improved performance as shown in Fig.~\ref{fig:baseline_results}. 

\begin{figure*}[!b]
    \begin{subfigure}[t]{0.3\textwidth}
      \includegraphics[width=\textwidth]{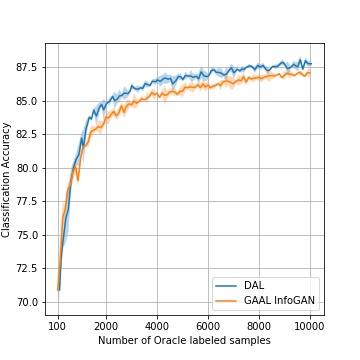}
      \caption{Effect of Automatic labeling}
      \label{fig:auto}
    \end{subfigure}
    \hfill
    \begin{subfigure}[t]{0.3\textwidth}
      \includegraphics[width=\textwidth]{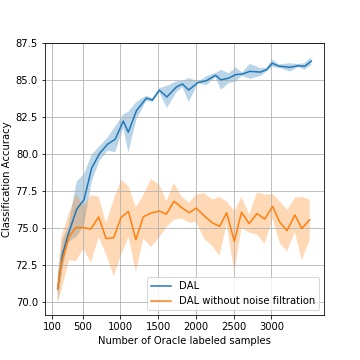}
      \caption{Effect of Label correction}
      \label{fig:correction}
    \end{subfigure}
    \hfill
    \begin{subfigure}[t]{0.3\textwidth}
      \includegraphics[width=\textwidth]{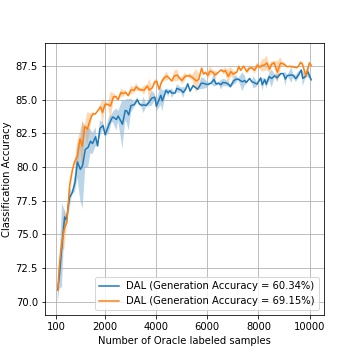}
      \caption{Effect of Disentanglement}
      \label{fig:disentanglement}
    \end{subfigure}
    
    \caption{Effect of various components in the performance of DAL}
  \end{figure*}
  
\section{Ablation Studies}
In this section, we perform various ablation studies to study the effectiveness of various components in our algorithm. Ablation studies performed in Subsections \ref{subsection:auto_label}, \ref{subsection:label_correct}, \ref{subsection:disentangle}, \ref{subsection:oracle_usage} are for Fashion MNIST dataset since the degree of disentanglement in Fashion MNIST is very less as discussed in the previous section and hence it would help us to analyze the worst case.

\subsection{Effect of Automatic labeling}\label{subsection:auto_label}
     To study the effect of automatic labeling involved in DAL, we compare our algorithm with GAAL which uses InfoGAN where InfoGAN is used to generate instances uniformly from all the classes and all instances are given to the oracle to label in each cycle. As shown in Fig.~\ref{fig:auto}, DAL can achieve better performance with a very less labeling budget compared to GAAL with InfoGAN. This makes it apparent that automatic labeling introduced in DAL with the help of disentanglement helps in the reduction of the labeling budget without degrading the performance of the active learner. 
     
\subsection{Effect of Label correction}\label{subsection:label_correct}
    We remove the label correction technique from DAL and study how it affects the performance of DAL. As illustrated in Fig.~\ref{fig:correction}, the accuracy of DAL in absence of label correction starts oscillating around 76\% as the active learning cycle proceeds. The degradation in performance could be attributed to the noisy instances that are added to dataset because of both the Active learner and the Generator agreeing on incorrect labels while the image is from a different class. Our technique alleviates this problem to a large extent which is pronounced by the performance of DAL as shown in Fig.~\ref{fig:correction}. Other noise-robust training methods such as Co-teaching introduced in \cite{NIPS2018_8072} or Bootstrapping as in \cite{43273} could also be incorporated to reduce the amount of noise in our system.
    
\subsection{Effect of Disentanglement}\label{subsection:disentangle}
    To study the effect of disentanglement on DAL,
    we consider two InfoGAN models with generation accuracy of 69.15\% and 60.34\% respectively. We compare the performance between DAL with each of these models. Fig.~\ref{fig:disentanglement} shows that, better the disentanglement, the lesser the number of human labeled datapoints required to achieve the fully supervised accuracy.
    
\begin{table*}[t]
  \centering
   \caption{Performance of Classifier trained on images generated by GAN trained on different number of labeled instances.}
    \label{tab:cgan}
   \begin{tabular}{l  l  l}
    \hline
    \textbf{GAN type} & \textbf{Number of labeled instances used to train GAN} & \textbf{Classifier Accuracy}\\
    \hline 
     CGAN & 1000 & 73.80\%\\
    CGAN & 5000 &  77.48\%  \\
    CGAN & 10000 & 76.09\% \\
    InfoGAN & \textbf{100} & \textbf{86.65\%}\\
     \hline
    \end{tabular}
\end{table*}

\begin{figure}[!t]
\begin{center}
\includegraphics[width=0.35\textwidth]{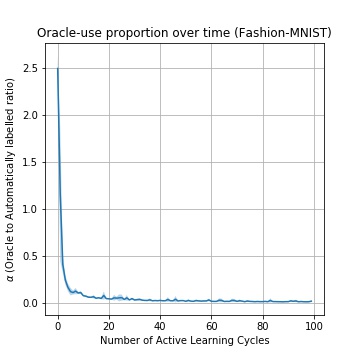}
\end{center}
  \caption{Ratio of Oracle labeled to Automatic labeled instances over time.}
\label{fig:oracle_usage}
\end{figure}

\subsection{Oracle usage over time}\label{subsection:oracle_usage}
    To study how often the oracle is queried over time, we plot Oracle-use proportion $\alpha$, the ratio of the number of oracle labeled to the number of automatically labeled instances versus active learning cycle. As illustrated in Fig.~\ref{fig:oracle_usage}, the oracle use proportion decreases as time progresses. As active learning progresses, most of the data points are automatically labeled and fewer points are given to oracle for labeling thus significantly alleviating the burden on manual annotation. The success of DAL could also be attributed to an increase in automatically labeling.
    
\subsection{Why is Conditional GAN not used in DAL?}
 Conditional GANs require supervision for training while InfoGAN learns to disentangle class categories in an unsupervised fashion. We performed experiments to analyse the discriminative power of classifiers trained on data from CGAN and InfoGAN. We trained multiple CGAN models with 1000, 5000 and 10000 labeled instances respectively from Fashion MNIST to generate 60000 instances from each model. We then trained classifiers using each of the three generated datasets and computed its accuracy. We also computed the same metric for a classifier trained on imaged generated from InfoGAN which is trained on 100 labeled instances and remaining unlabeled instances. The results are reported in Table~\ref{tab:cgan}. The experimental analysis clearly exhibits the ineffectiveness of CGANs when labeled instances are very less.
 
 \begin{figure}[!t]
\begin{center}
\includegraphics[width=0.35\textwidth]{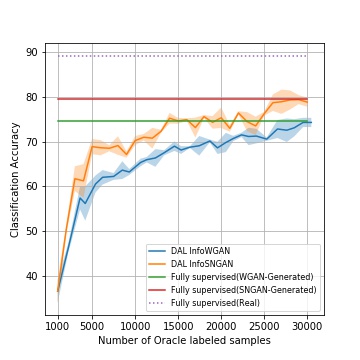}
\end{center}
  \caption{Performance of DAL using InfoSNGAN compared to that of InfoWGAN on \textit{CIFAR-10} dataset.}
\label{fig:sngan_cifar}
\end{figure}
 
\subsection{Effect of Improved GAN Architecture}
We trained an InfoGAN combined with SNGAN \cite{sngan} (InfoSNGAN) on \textit{CIFAR-10} to generate images with better quality compared to InfoWGAN. We used InfoSNGAN in DAL to compare it against DAL with InfoWGAN on \textit{CIFAR-10}. Fig.~\ref{fig:sngan_cifar} shows that performance of DAL with InfoSNGAN is much better compared to that of DAL with InfoWGAN which makes it evident that better performance of DAL can be attributed to usage of improved GAN architectures and training schemes.

\section{Conclusion And Future Work}
We proposed a novel approach for active learning by integrating the concept of disentanglement along with GAN based active learning techniques and a simple feedback mechanism from the active learner. Using right heuristics as shown in \cite{DBLP:conf/icassp/BhattaraiBBK20}, \cite{gen_data} would help us to improve the performance of classifiers trained on synthesized images. Models that exhibit better disentanglement would also reduce the labeling budget in our algorithm to a large extent as demonstrated in ablation studies. Currently, we use a pre-trained InfoGAN based on which the active learner evolves as the active learning cycle proceeds. We would like to explore the idea of jointly training the InfoGAN and active learner as in \cite{BGAD} which seems to be an interesting direction. Extension of our approach to Pool based methods and Augmentation based active learning methods \cite{DBLP:journals/corr/abs-1904-11643} could also be taken up as future work.

 \bibliographystyle{IEEEtran}
\bibliography{DAL}

\end{document}